# Selection of Most Appropriate Backpropagation Training Algorithm in Data Pattern Recognition

Hindayati Mustafidah [#1], Sri Hartati [#2], Retantyo Wardoyo [#3], Agus Harjoko [#4]

*[1]Engineering Informatics - Muhammadiyah University of Purwokerto*
*Purwokerto, Indonesia*
*[234] Computer Science – FMIPA - Gadjah Mada University*
*Yogyakarta, Indonesia*

*Abstract*—There are several training algorithms for back propagation method in neural network. Not all of these algorithms have the same accuracy level demonstrated through the percentage level of suitability in recognizing patterns in the data. In this research tested 12 training algorithms specifically in recognize data patterns of test validity. The basic network parameters used are the maximum allowable epoch = 1000, target error = $10^{-3}$, and learning rate = 0.05. Of the twelve training algorithms each performed 20 times looping. The test results obtained that the percentage rate of the great match is trainlm algorithm with alpha 5% have adequate levels of suitability of 87.5% at the level of significance of 0.000. This means the most appropriate training algorithm in recognizing the the data pattern of test validity is the trainlm algorithm.

*Keywords*—validity, appropriate, training algorithms, data pattern recognition.

## I. INTRODUCTION

One of the computer technology development in the creation of the current approach in resolving an issue is soft computing [1]. Soft Computing is part of the intelligent system that is a model approach to computing by imitating human reason and has the ability to reason and learn in an environment that is filled with uncertainty and inaccuracy. Soft Computing consists of several main components constituent that is fuzzy systems, neural network, evolutionary algorithms, and probabilistic reasoning. Performance of soft computing in solving problems in various areas of life has proven its benefits. One of the methods used in this study is an Artificial Neural Network (ANN), which is the ideal solution for problems that can't be easily formulated using an algorithm [2]. In the neural network there are several learning algorithms. Based on how it works, learning algorithm is divided into two parts with adaptation-based and training-based learning. The adaptation-based learning algorithm consists of two algorithms namely learngd and learngdm, while the training-based learning algorithm consisting of twelve algorithms, they are traingd, traingdm, traingdx, trainrp, traingda, traincgf, traincgp, traincgb, trainscg, trainbfg, trainoss, and trainlm [3]. Training-based learning algorithm is often referred to as a training algorithm.

The use of training algorithm in neural network has been done in solving, for example, predicts student learning achievement at Informatics Engineering in Muhammadiyah University of Purwokerto that is based on the grades in the subjects to be tested in the national exam as in High School [4]. The training algorithm used is traingd and produces error rate of 0.0664179 of target error 0.05. Seen that the resulting error rates in excess of a expected target error. It shows that it needs to review and study the learning algorithm that held the most optimal in problem solving. Some studies have also been carried out to determine the most optimal algorithm that gives the smallest error i.e. comparing the optimum level of training algorithms traingd, traingdm, learngd, and learngdm in case of student learning achievement predictions at Informatics Engineering Muhammadiyah University of Purwokerto [5]. Based on the results of the statistical tests of the four algorithms with the alpha level 5%, obtained the value significance of 0,632. It means that the four algorithms have no differences in optimum level. In other words, the four training algorithms have the same proper level. Based on this case, it needs to be done further testing on other training algorithms to obtain reference the most appropriate one.

On this research was performed testing against the twelve training algorithms to obtain the most appropriate algorithm in data pattern recognizing. Appropriateness is shown by the percentage of matches in recognizing patterns of data i.e. data patterns of training and test data pattern where the test data used is similar to the training data. A case study used is of data validity of test questions based on the percentage category of test item in Bloom revised taxonomy developed by Anderson-Krathwohl. This Data is used in the this research by reason of the training algorithms obtained will be used in the determination of the level of the validity of the question test with a different case so that will be obtained the right predictions of reserved test questions validity.

Validity is a measure that shows the levels of instrument validity. A test is said to have validity if the result corresponds to criterion, in the sense of having equality between the test results and criteria [6]. A test called valid if the test were able to precisely measure what is measured. If the data that is retrieved from a valid instrument, it can be said that the instrument is valid as it can give a description of the data correctly in accordance with the fact or circumstance indeed. In doing the analysis of validity about commonly used correlation product moment formula [7]. A test is said to have the validity of the construction when the details matter that build these tests measure every aspect of thinking as mentioned in the learning objectives. Construction in this sense is psychological invention itemize the contents of several aspects such as: memory/knowledge, comprehension, application, analysis, synthesis, and evaluation better known as Bloom's taxonomy [8]. The six aspects by [9] developed as six cognitive processes dimensions, namely: remembering (C1), understanding (C2), applying (C3), analyzing (C4), evaluating (C5), and creating (C6) that known a revision of Bloom's Taxonomy by Anderson-Krathwohl. The validity of the construction can be known with detailing and pair each test item with every aspect of the learning objectives.

## II. RESEARCH METHODS

The data used in this research are primary and secondary data. Primary data are student's scores and test questions at National Exam in 2009 of the province of Yogyakarta and Bengkulu. The primary data retrieved from the PUSPENDIK (Centre for Educational Assessment) in Jakarta, while the secondary data were retrieved from the data analysis of the research about test analysis at 2003, 2004, 2005 in Banyumas Regency and its surroundings ([10], [11], [12], [13], [14], [15], [16], [17], [18], [19]). The research data were presented in Table I.

TABLE I
RESEARCH DATA

| Percentage | | | | | | Validity |
|---|---|---|---|---|---|---|
| C1 | C2 | C3 | C4 | C5 | C6 | |
| 20 | 35 | 0 | 45 | 0 | 0 | 0.351 |
| 0 | 15 | 29 | 6 | 0 | 0 | 0.308 |
| 0 | 15 | 29 | 6 | 0 | 0 | 0.376 |
| 18.75 | 12.5 | 59.38 | 9.38 | 0 | 0 | 0.232 |
| 12.5 | 6.25 | 71.88 | 9.38 | 0 | 0 | 0.205 |
| 12.9 | 9.68 | 67.74 | 9.68 | 0 | 0 | 0.186 |
| 16.67 | 56.67 | 23.33 | 3.33 | 0 | 0 | 0.752 |
| 14.29 | 62.86 | 20 | 2.86 | 0 | 0 | 0.296 |
| 17.14 | 57.14 | 17.14 | 8.57 | 0 | 0 | 0.345 |
| 50 | 30 | 10 | 10 | 0 | 0 | 0.361 |
| 50 | 30 | 10 | 10 | 0 | 0 | 0.232 |
| 48 | 42 | 6 | 4 | 0 | 0 | 0.071 |
| 48 | 42 | 6 | 4 | 0 | 0 | 0.175 |
| 60 | 36.67 | 3.33 | 0 | 0 | 0 | 0.218 |
| 28.89 | 37.78 | 26.67 | 6.67 | 0 | 0 | 0.184 |
| 28.89 | 37.78 | 26.67 | 6.67 | 0 | 0 | 0.221 |
| 28.89 | 37.78 | 26.67 | 6.67 | 0 | 0 | 0.155 |
| 40 | 38 | 8 | 14 | 0 | 0 | 0.162 |
| 40 | 38 | 8 | 14 | 0 | 0 | 0.201 |
| ... | | | | | | |
| 0 | 0 | 45 | 52.5 | 0 | 2.5 | 0.458 |

This research uses mixed method, namely qualitative and quantitative research. The qualitative research carried out by developing a program code for processing data using the twelve algorithms (Fig. 1), whereas quantitative research conducted by statistical testing against the accuracy of the twelve algorithms in recognizing data patterns shown by the percentage of matches between the results of network simulation data and the input data. These algorithms are traingd, traingdm, traingdx, trainrp, traingda, traincgf, traincgp, traincgb, trainscg, trainbfg, trainoss, and trainlm. The output data of any algorithm is done by looping 20 times.

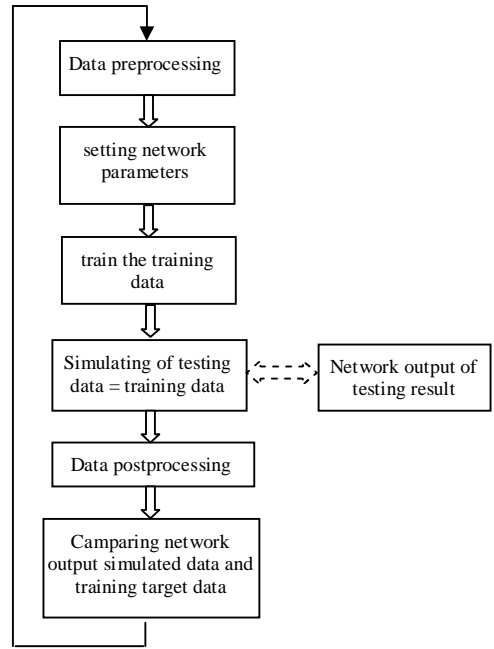

Fig. 1 The process of training and testing data

Statistical testing is done using SPSS software to determine the most appropriate algorithm using one-way analysis of variance (One-way ANOVA) [20]. The hypotheses used are:

$H_0$: the variants of the twelve training algorithms are the same
$H_1$: the variants of the twelve training algorithms are not the same.

The rejection of $H_0$ is performed if the value significance obtained is less than α (in this case α = 0.05 or 5%).

## III. RESULTS AND DISCUSSION

### A. The Structure of the Network

The network structure is shown in Fig. 2.

### B. The Network Parameters

Parameters of the network includes the learning rate (LR) = 0.05, the maximum amount of epoch = 1000, and the level of error = 0.001 ($10^{-3}$).

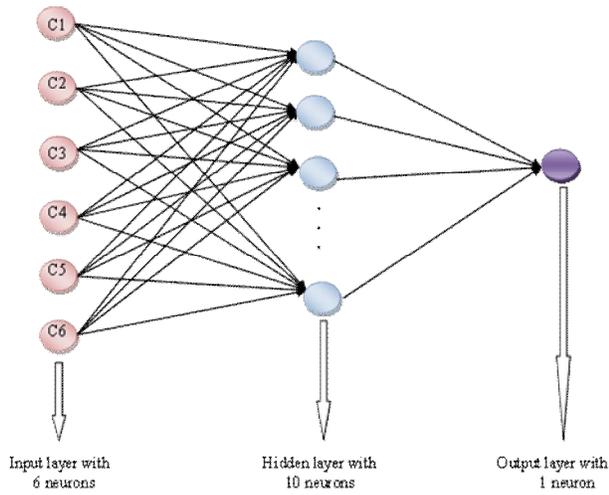

Fig. 2 Neural network structure with 3 layers

*C. Testing*

ANOVA test results obtained significance of 0.000, it means that the value is less than α, so $H_0$ is rejected. In other word, the twelve algorithms are not homogeneous (have not the same variant) (Table II).

TABLE II
ANOVA TEST RESULTS OF THE 12 TRAINING ALGORITHMS USING SPSS

Matched reached in percentages

|  | Sum of Squares | df | Mean Square | F | Sig. |
|---|---|---|---|---|---|
| **Between Groups** | 25020.286 | 11 | 2274.571 | 129.398 | 0.000 |
| **Within Groups** | 4007.812 | 228 | 17.578 |  |  |
| **Total** | 29028.099 | 239 |  |  |  |

To learn more the algorithms which have differences level of appropriateness, performed using a Duncan POSTHOC method with results as shown in the Table III.

Based on the Duncan method, it shows that traincgf, trainscg, traincgb, and trainlm algorithms are on the lowest subset with significance 0.086. This means that the four training algorithms have the same appropriateness level at α = 5%. To get the most appropriate algorithm of four the training algorithms, further testing needs to be done and the results as presented in Table IV.

From the second test result, significance of 0.003 is obtained which means that all four training algorithms have inequality on the appropriateness level. Then the testing is done using Duncan method that is indicated as in Table V.

It is seen that in this test was obtained similarities at two algorithms, namely traincgb and trainlm on the significance of 0.053. Because it has not yet obtained the most appropriate algorithm, testing was continued with an average difference of testing against both algorithms using independent samples t-test (Table VI).

TABLE III
POSTHOC TEST OF DUNCAN MULTIPLE COMPARISONS ANALYSIS FOR 12 ALGORITHMS USING SPSS

Matched reached in percentages

| Analysis | Training Algorithms | N | Subset for alpha = 0.05 | | | | |
|---|---|---|---|---|---|---|---|
|  |  |  | 1 | 2 | 3 | 4 | 5 |
| Duncan[a] | traingda | 20 | 58.125 |  |  |  |  |
|  | traingd | 20 |  | 64.125 |  |  |  |
|  | traingdm | 20 |  | 65.750 |  |  |  |
|  | traingdx | 20 |  | 66.375 |  |  |  |
|  | trainrp | 20 |  |  | 80.500 |  |  |
|  | trainoss | 20 |  |  |  | 84.000 |  |
|  | traincgp | 20 |  |  |  | 84.121 |  |
|  | trainbfg | 20 |  |  |  | 84.375 |  |
|  | traincgf | 20 |  |  |  | 85.000 | 85.000 |
|  | trainscg | 20 |  |  |  | 85.375 | 85.375 |
|  | traincgb | 20 |  |  |  | 86.125 | 86.125 |
|  | trainlm | 20 |  |  |  |  | 87.500 |
|  | Sig. |  | 1.000 | 0.110 | 1.000 | 0.166 | 0.086 |

Means for groups in homogeneous subsets are displayed.
a. Uses Harmonic Mean Sample Size = 20.000.

TABLE IV
ANOVA TEST RESULTS OF 4 TRAINING ALGORITHMS

Matched reached in percentages

|  | Sum of Squares | df | Mean Square | F | Sig. |
|---|---|---|---|---|---|
| **Between Groups** | 73.125 | 3 | 24.375 | 4.982 | 0.003 |
| **Within Groups** | 371.875 | 76 | 4.893 |  |  |
| **Total** | 445.000 | 79 |  |  |  |

TABLE V
POSTHOC TEST OF DUNCAN MULTIPLE COMPARISONS ANALYSIS FOR 4 ALGORITHMS

Duncan

| Algorithm traincgf, trainscg, traincgb, trainlm | N | Subset for alpha = 0.05 | |
|---|---|---|---|
|  |  | *1* | *2* |
| traincgf | 20 | 85.000 |  |
| trainscg | 20 | 85.375 |  |
| traincgb | 20 | 86.125 | 86.125 |
| trainlm | 20 |  | 87.500 |
| Sig. |  | 0.133 | 0.053 |

Means for groups in homogeneous subsets are displayed.

Based on Table VI seen that with 95% confidence interval and significance level of α = 0.000 (< 5%), the traincgb algorithm is in the negative area which means that it has smaller appropriateness level than trainlm.

TABLE VI

THE TEST RESULTS OF THE INDEPENDENT SAMPLE T-TEST

Independent Samples Test

| | | Levene's Test for Equality of Variances | | t-test for Equality of Means | | | | | 95% Confidence Interval of the Difference | |
|---|---|---|---|---|---|---|---|---|---|---|
| | | *F* | Sig. | *t* | *df* | Sig. (2-tailed) | Mean Difference | Std. Error Difference | Lower | Upper |
| **Matched of traincgb & trainlm** | Equal variances assumed | 73.962 | 0.000 | -3.240 | 38 | 0.002 | -1.375000 | 0.424380 | -2.234113 | -0.515887 |
| | Equal variances not assumed | | | -3.240 | 19.000 | 0.004 | -1.375000 | 0.424380 | -2.263238 | -0.486762 |

IV. CONCLUSIONS

The 12 training algorithms in neural network have been tested, namely traingd, traingdm, traingdx, trainrp, traingda, traincgf, traincgp, traincgb, trainscg, trainbfg, trainoss, and trainlm in the appropriateness of recognizing the data pattern. The network parameters include maximum epoch = 1000, the learning rate = 0.05, and target error = 0.001. The test results obtained that at the confidence interval of 95% the trainlm algorithm is the algorithm that is most appropriate in recognizing the data pattern with an average level of appropriateness of 87.5%. It is advisable to test the training algorithms in recognizing data patterns in the other case of categories of test quality namely reliability, test difference power, and difficulty level reserved.